\documentclass[conference]{IEEEtran}
\usepackage[latin9]{inputenc}
\usepackage{array}
\usepackage{url}
\usepackage{multirow}
\usepackage{amsmath}
\usepackage{amssymb}
\usepackage{graphicx}

\makeatletter

\providecommand{\tabularnewline}{\\}

\IEEEoverridecommandlockouts
\usepackage{cite}
\usepackage{amsfonts}\usepackage{algorithmic}
\usepackage{textcomp}
\usepackage{xcolor}
\def\BibTeX{{\rm B\kern-.05em{\sc i\kern-.025em b}\kern-.08em
    T\kern-.1667em\lower.7ex\hbox{E}\kern-.125emX}}

\makeatother

\begin{document}
\title{On guiding video object segmentation\\
}
\author{\IEEEauthorblockN{Diego Ortego} \IEEEauthorblockA{Insight Centre for Data Analytics \\
Dublin City University (DCU) \\
Dublin (Ireland) \\
diego.ortego@insight-centre.org} \and \IEEEauthorblockN{Kevin McGuinness} \IEEEauthorblockA{Insight Centre for Data Analytics \\
Dublin City University (DCU) \\
Dublin (Ireland) \\
kevin.mcguinness@insight-centre.org} \and \IEEEauthorblockN{Juan C. SanMiguel} \IEEEauthorblockA{VPULab \\
Universidad Autónoma de Madrid (UAM) \\
Madrid (Spain) \\
juancarlos.sanmiguel@uam.es} \and \IEEEauthorblockN{Eric Arazo} \IEEEauthorblockA{Insight Centre for Data Analytics \\
Dublin City University (DCU) \\
Dublin (Ireland) \\
eric.arazo@insight-centre.org} \and \IEEEauthorblockN{José M.~Martínez} \IEEEauthorblockA{VPULab \\
Universidad Autónoma de Madrid (UAM) \\
Madrid (Spain) \\
josem.martinez@uam.es} \and \IEEEauthorblockN{Noel E. O'Connor} \IEEEauthorblockA{Insight Centre for Data Analytics \\
Dublin City University (DCU) \\
Dublin (Ireland) \\
noel.oconnor@insight-centre.org}}
\maketitle
\begin{abstract}
This paper presents a novel approach for segmenting moving objects
in unconstrained environments using guided convolutional neural networks.
This guiding process relies on foreground masks from independent algorithms
(i.e. state-of-the-art algorithms) to implement an attention mechanism
that incorporates the spatial location of foreground and background
to compute their separated representations. Our approach initially
extracts two kinds of features for each frame using colour and optical
flow information. Such features are combined following a multiplicative
scheme to benefit from their complementarity. These unified colour
and motion features are later processed to obtain the separated foreground
and background representations. Then, both independent representations
are concatenated and decoded to perform foreground segmentation. Experiments
conducted on the challenging DAVIS 2016 dataset demonstrate that our
guided representations not only outperform non-guided, but also recent
and top-performing video object segmentation algorithms.
\end{abstract}


\section{Introduction}

\IEEEPARstart{S}{egmenting} an image into regions is key for identifying
objects of interest. For example, image segmentation \cite{2012_TPAMI_SLIC}
clusters image pixels with common properties (e.g. colour, textures,
etc), while semantic segmentation \cite{2017_CVPR_PSPNet} categorises
each pixel into a set of predefined classes. Foreground segmentation
is a particular case of semantic segmentation with two categories:
foreground and background. The former contains the objects of interest
in an image, which may correspond to salient objects \cite{2015_TIPSaliencyQ}\cite{2017_SPL_SpixFGextract},
generic objects \cite{2012_TPAMIObjectness}\cite{2017_arXiv_Objectness},
moving objects \cite{2017_CVPR_ARP}, spatio-temporal relevant patterns
\cite{2011_ICCV_KeySeg} or even weak labels \cite{2017_CVPR_SelfPacedVOS}.
The latter consists on the non-relevant data, being usually the static
scene objects. Moreover, foreground segmentation in unconstrained
environments is known as video object segmentation (VOS) and faces
many challenges related to camera motion, shape deformations, occlusions
or motion blur \cite{2014_BMVC_Irani}.

VOS has become an active research area, as demonstrated by the widespread
use of the DAVIS benchmark \cite{2016_CVPR_Davis}. Existing algorithms
are supervised, semi-supervised and unsupervised. Supervised approaches
employ frame-by-frame human intervention \cite{2017_arXiv_InteractiveVOS}
whereas semi-supervised ones only require initialization (e.g. annotations
of the objects to segment in the first frame) \cite{2014_SPL_VOcutout}\cite{2017_CVPR_CTN}.
Conversely, unsupervised VOS does not involve human intervention,
requiring the automatic detection of relevant moving objects \cite{2017_CVPR_FSEG}.
\begin{figure*}[t]
\centering{}\includegraphics[width=1\textwidth]{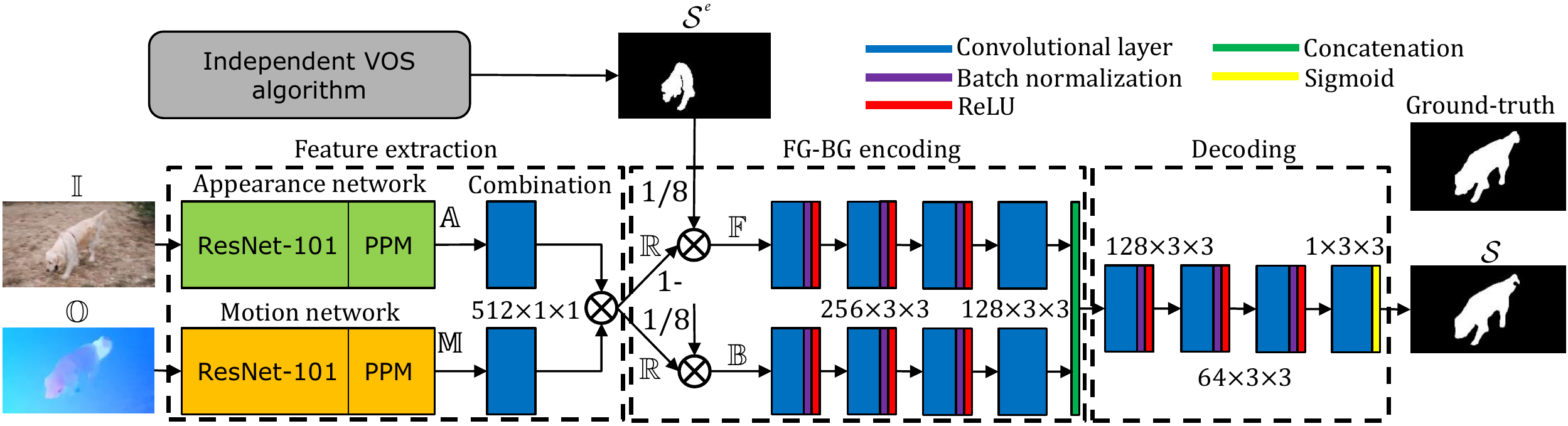}\caption{\label{fig:Arquitecture}Overview of the proposed architecture. Video
object segmentation is performed making use of appearance and motion
convolutional representations computed using features extracted after
the Pyramid Pooling Module (PPM) of PSPNet \cite{2017_CVPR_PSPNet}.
Then, such representations are decoded with the guide of an attention
mechanism provided by the foreground segmentation mask $\mathcal{S}^{e}$
computed using an independent algorithm. Key: FG (foreground). BG
(background).}
\end{figure*}

Currently deep learning is significantly advancing computer vision
performance \cite{2015_Nature_DeepLearning} such as for VOS, where
state-of-the-art approaches employ convolutional neural networks \cite{2017_CVPR_ARP}\cite{2017_CVPR_MSK}.
Performance improvement can be achieved by increasing the complexity
of the network \cite{2012_NIPS_AlexNet}\cite{2016_CVPR_ResNet},
but also by learning better models without requiring new architectures.
For instance, loss function variations \cite{2017_ICCV_FocalLoss},
transfer learning \cite{2016_TPAMI_FactorsTransf}, data augmentation
\cite{2014_BMVC_DataAugmentationeltit} or applying spatial attention
\cite{2017_BMVC_ReverseAttention} are techniques widely explored.
In particular, spatial attention can be used to highlight activations
in feature maps of the network, thus enabling training of more accurate
models. In \cite{2017_BMVC_ReverseAttention}, attention is extracted
from convolutional features from a semantic segmentation network to
promote those activations that do not respond to several classes,
but to only one. Also, \cite{2017_ICCV_HydraNet} generates attention
maps responding to visual patterns in different scales and levels
from convolutional features, thus capturing different semantic patterns
such as body parts, objects, or background that improve pedestrian
attribute recognition. Furthermore, \cite{2018_CVPR_TrackingAttention}
uses three attentions maps (general, residual and channel attentions)
extracted from convolutional features to weight the cross-correlation
of a fully-convolutional siamese tracker that adapts the offline learned
model to the online tracked targets. Additionally, content-based image
retrieval \cite{2018_CBMI_Saliency} can be enhanced by using visual
attention models to weight the contribution of the activations from
different spatial regions.

This paper proposes a novel approach to improving the performance
of an unsupervised VOS algorithm by using an independent foreground
segmentation (i.e. the mask from an existing algorithm in the literature)
to guide the segmentation process by focusing on relevant activations.
First, given a video frame and its associated optical flow, two networks
compute appearance and motion feature maps. Second, both feature maps
are unified to exploit their complementarity. Third, the foreground
mask of an independent algorithm is used to encode foreground and
background information. Finally, both representations are concatenated
and decoded to produce a foreground mask. We validate the proposed
approach in the recent DAVIS 2016 dataset \cite{2016_CVPR_Davis},
demonstrating the improvements achieved by our approach, whose novelty
lies in the use of an independent foreground mask to separate foreground
and background representations and learn a better foreground segmentation
model.

The remainder of this paper is organized as follows: Section \ref{sec:Algorithm-overview}
overviews the proposed approach whereas Section \ref{sec:Algorithm-description}
and \ref{sec:Experimental-work} describe the proposed algorithm and
the experiments performed. Finally, Section \ref{sec:Conclusions}
concludes this paper.

\section{Algorithm overview\label{sec:Algorithm-overview}}

\noindent VOS can be formulated as a pixel-wise labelling of each
video frame $\mathbb{I}$ as either foreground (1) or background (0),
thus generating a foreground segmentation mask $\mathcal{S}$. We
propose an approach based on convolutional neural networks (CNNs)
that uses a frame $\mathbb{I}$ together with its corresponding optical
flow $\mathbb{O}$ and the foreground mask of an independent algorithm
$\mathcal{S}^{e}$ (from now on, the independent foreground mask),
to compute a foreground segmentation $\mathcal{S}$ (see Figure \ref{fig:Arquitecture}).
Our CNN-VOS approach starts with an appearance network computing a
convolutional feature map $\mathbb{A}$ from a video frame $\mathbb{I}$
and a motion network generating a convolutional feature map $\mathbb{M}$
from the optical flow $\mathbb{O}$ associated to $\mathbb{I}$. Both
feature maps are then combined by element-wise multiplication to obtain
a unique representation $\mathbb{R}$ comprising both appearance and
motion information. Subsequently, the independent foreground mask
$\mathcal{S}^{e}$ is used to weight $\mathbb{R}$ and separately
encode foreground $\mathbb{F}$ and background $\mathbb{B}$ information
based on the previously computed feature maps, which are processed
by two independent networks. Finally, the output of the two sub-networks
is concatenated and decoded with a convolutional network to produce
the foreground mask $\mathcal{S}$.

\section{Algorithm description\label{sec:Algorithm-description}}

\subsection{Appearance network}

\noindent The appearance network learns a model $\mathbb{A}$ from
spatial distribution of the colour information of a video frame $\mathbb{I}$.
In particular, we use PSPNet \cite{2017_CVPR_PSPNet}, a fully-convolutional
neural network for semantic segmentation which relies on ResNet \cite{2016_CVPR_ResNet}
followed by a hierarchical context module (Pyramid Pooling Module)
that harvest information from different scales. This module processes
different downsamplings of ResNet features and later upsamples them
for concatenating all features to form an improved representation
that includes the original ResNet features. In particular, we use
ResNet-101 to obtain $\mathbb{A}$ by extracting the 512 feature maps
obtained after the convolutional layer that follows the Pyramid Pooling
Module, which contains both local and global representations.

\subsection{Motion network}

In contrast to the appearance network, the motion network obtains
a model $\mathbb{M}$ again making use of PSPNet, but in this case
from the optical flow $\mathbb{O}$. In particular, we convert the
optical flow vectors \cite{2009_OF_Liu} into a 3-channel colour-coded
optical flow image $\mathbb{O}$ \cite{2017_CVPR_FSEG} and train
PSPNet to produce a foreground segmentation from the colour-coded
optical flow $\mathbb{O}$. In order to get the model $\mathbb{M}$,
we select the 512 feature maps at the same layer as done in the appearance
network.

\subsection{Unified representation}

\noindent The intuition behind using two networks with independent
inputs (appearance and motion) is to benefit from the complementary
of appearance and motion exhibited by moving objects \cite{2017_ICCV_GRU}.
Therefore, we exploit such complementarity by combining both feature
maps following a multiplicative fusion (see the Combination stage
in Figure \ref{fig:Arquitecture}) similarly to \cite{2016_WACV_InfCombination},
where the authors multiply different sources of CNNs (appearance and
motion) to enable the amplification or suppression of feature activations
based on their agreement. In our approach, this fusion consists of
a $1\times1$ convolution applied to both the appearance features
$\mathbb{A}$ and the motion features $\mathbb{M}$ followed by an
element-wise multiplication of both sets of feature maps (which have
the same dimensionality) to produce the unified encoding $\mathbb{R}$.
Applying the $1\times1$ convolutions helps to control the dimensionality
while learning how to combine the feature maps before their multiplicative
combination.

\subsection{Foreground and background encoding}

Unlike many VOS literature which jointly process all features \cite{2017_CVPR_FSEG},
we introduce an attention mechanism that splits the feature maps into
foreground and background to better guide the learning process and
obtain a better VOS model (see FG-BG encoding stage in Figure \ref{fig:Arquitecture}).
In particular, we use an foreground mask $\mathcal{S}^{e}$ (obtained
from a state-of-the-art algorithm) which is downsampled to match the
size of $\mathbb{R}$. Then, we split $\mathbb{R}$ into foreground
$\mathbb{F}=\mathbb{R}\times\mathbb{S}^{e}$ and background $\mathbb{B}=\mathcal{\mathbb{R}}\times\left(1-\mathbb{S}^{e}\right)$
representations according to the guidance provided by the independent
foreground $\mathcal{S}^{e}$. On the one hand, multiplying the feature
maps $\mathbb{R}$ by the independent foreground $\mathcal{S}^{e}$
helps focusing on important spatial areas of the features by zeroing
responses associated to background areas in $\mathbb{F}$, thus implementing
an attention mechanism. On the other hand, maintaining information
from the background regions through $\mathbb{B}$ helps in the segmentation
process, which may be useful as it assures a background representation
is maintained, especially when errors in $\mathcal{S}^{e}$ lead to
the suppression of important foreground responses. Subsequently, $\mathbb{F}$
and $\mathbb{B}$ are individually processed by four convolutional
layers to separately model foreground and background representations
before concatenating them as presented in the \emph{FG-BG encoding}
stage in Figure \ref{fig:Arquitecture}. Finally, this set of concatenated
foreground and background representations contain a high-level joint
encoding of appearance and motion that is fed to four additional convolutional
layers for decoding in order to compute the final prediction (see
Decoding stage in Figure \ref{fig:Arquitecture}), i.e. the foreground
segmentation mask $\mathcal{S}$.

\subsection{Architecture details}

Our architecture is the fully-convolutional network shown in Figure
\ref{fig:Arquitecture}. After the appearance and motion networks
(both using the PSPNet architecture), there are two convolutional
layers that use 512 $1\times1$ kernels to process the feature maps
$\mathbb{A}$ and $\mathbb{M}$ (both have 1/8 the original image
resolution). Then, the multiplication of $\mathcal{S}^{e}$ is preceded
by a downsampling performed through average pooling with stride 8.
Regarding the processing of $\mathbb{F}$ and $\mathbb{B}$: before
concatenating both we use four convolutional layers, three with depth
256 followed by batch normalization and ReLU and the final one with
depth 128 to reduce the number of feature maps. As it is well known
that reducing the feature map resolution may result in coarse predictions
\cite{2017_TPAMI_FCN}, we use dilated convolutions (1, 2, 4 and 8
depths) to aggregate context \cite{2015_arXiv_DilatedConv} while
preserving the already reduced resolution (1/8). Finally, we apply
four convolutional layers with dilated convolutions (128, 64, 64 and
1 depths). We use $3\times3$ kernels throughout (unless otherwise
stated).

\section{Experimental work\label{sec:Experimental-work}}

\subsection{Datasets and metrics}

We evaluate the proposed approach on the DAVIS 2016 dataset \cite{2016_CVPR_Davis}
which covers many challenges of unconstrained VOS. We use all available
test videos (20) whose length ranges from 40 to 100 frames with $854\times480$
resolution. A unique foreground object is annotated in each frame.
We use three standard performance measures \cite{2016_CVPR_Davis}
to account for region similarity (Jaccard index $\mathcal{J}$, i.e.
intersection-over-union ratio between the segmented foreground mask
and the ground-truth mask), for contour accuracy (F-score $\mathcal{F}$
between contour pixels of the segmented and the ground-truth masks)
and for temporal stability $\mathcal{T}$ of foreground masks (by
associating temporal smooth and precise transformations to good foreground
segmentations). These measures are computed frame-by-frame and then
averaged to compute a sequence-level score, which are also averaged
to get dataset-level performance. Note that we do not consider DAVIS
2017 dataset as it is semi-supervised whereas our approach is unsupervised.

\subsection{Implementation details\label{subsec:Implementation-details}}

The training procedure consists of three stages. First, we train the
appearance network to perform semantic segmentation on the PASCAL
VOC 2012 dataset \cite{2010_IJCVPASCAL} using a total of 10582 training
images. Second, we train the motion network to perform foreground
segmentation based on optical flow data \cite{2009_OF_Liu} using
the annotations provided by \cite{2017_CVPR_FSEG} for 84929 frames
of ImageNet-Video dataset \cite{2015_ImageNet}. Third, we freeze
both the appearance and the motion networks and train the rest of
the network on 22 of the 30 training video sequences of DAVIS 2016
dataset \cite{2016_CVPR_Davis} (we use 8 sequences for validation).
For this last step, masks of an independent algorithm are needed,
so we use available foreground masks in DAVIS 2016\footnote{\url{https://davischallenge.org/davis2016/soa_compare.html}}
(the 13 algorithms evaluated in \cite{2016_CVPR_Davis} and the algorithm
proposed in \cite{2016_CVPR_BVS}). We train the appearance and the
motion networks for 30k and 90k iterations, respectively, using the
``poly\textquotedblright{} learning rate policy described in \cite{2017_CVPR_PSPNet}.
However, for the third step we train the network for 20 epochs reducing
the learning rate by ten each five epochs (starting with the value
0.1) and we select the best model using the performance in the validation
set. We use batch size 8, data augmentation (random Gaussian blur,
sized crops, rotations and horizontal flips), cross-entropy loss and
\textit{Kaiming} weight initialization \cite{2015_ICCV_NetInit} in
all training steps.

\subsection{Evaluation\label{subsec:Evaluation}}

\begin{table}[t]
\centering{}\caption{\label{tab:PerformanceDAVIS}Comparative results for DAVIS 2016. ({*})
indicates the proposed approach using the corresponding foreground
mask. $\uparrow\left(\downarrow\right)$ means higher (lower) is better.
Bold denotes better performance for our approach.}
\begin{tabular}{|l|c|c|c|c|c|c|}
\hline 
\multirow{2}{*}{Algorithms} & \multicolumn{2}{c|}{$\mathcal{J}\left(\uparrow\right)$} & \multicolumn{2}{c|}{$\mathcal{F\left(\uparrow\right)}$} & \multicolumn{2}{c|}{$\mathcal{T\left(\downarrow\right)}$}\tabularnewline
\cline{2-7} \cline{3-7} \cline{4-7} \cline{5-7} \cline{6-7} \cline{7-7} 
 & mean & std. & mean & std. & mean & std.\tabularnewline
\hline 
\hline 
ARP & .7609 & .1125 & .7051 & .1111 & .5363 & .3446\tabularnewline
\hline 
ARP{*} & \textbf{.8069} & \textbf{.0631} & \textbf{.8121} & \textbf{.0799} & \textbf{.3930} & \textbf{.2675}\tabularnewline
\hline 
CTN & .7304 & .1048 & .6886 & .1147 & .3682 & .2176\tabularnewline
\hline 
CTN{*} & \textbf{.7933} & \textbf{.0627} & \textbf{.7996} & \textbf{.0775} & .3969 & 2762\tabularnewline
\hline 
FSEG & .7068 & .0804 & .6524 & .1036 & .4456 & .2809\tabularnewline
\hline 
FSEG{*} & \textbf{.7834} & \textbf{.0673} & \textbf{.7950} & \textbf{.0773} & \textbf{.3938} & \textbf{.2566}\tabularnewline
\hline 
MSK & .7955 & .0817 & .7519 & .0929 & .3226 & .1978\tabularnewline
\hline 
MSK{*} & \textbf{.8012} & \textbf{.0718} & \textbf{.8097} & \textbf{.0821} & .3805 & .2660\tabularnewline
\hline 
OFL & .6738 & .1231 & .6279 & .1330 & .3608 & .2123\tabularnewline
\hline 
OFL{*} & \textbf{.7844} & \textbf{.0764} & \textbf{.7926} & \textbf{.0933} & .3898 & .2805\tabularnewline
\hline 
VPN & .6984 & .0902 & .6513 & .1002 & .4923 & .2630\tabularnewline
\hline 
VPN{*} & \textbf{.7765} & \textbf{.0709} & \textbf{.7920} & \textbf{.0792} & \textbf{.3918} & \textbf{.2551}\tabularnewline
\hline 
\end{tabular}
\end{table}

\begin{figure}[t]
\centering{}\includegraphics[width=0.99\columnwidth]{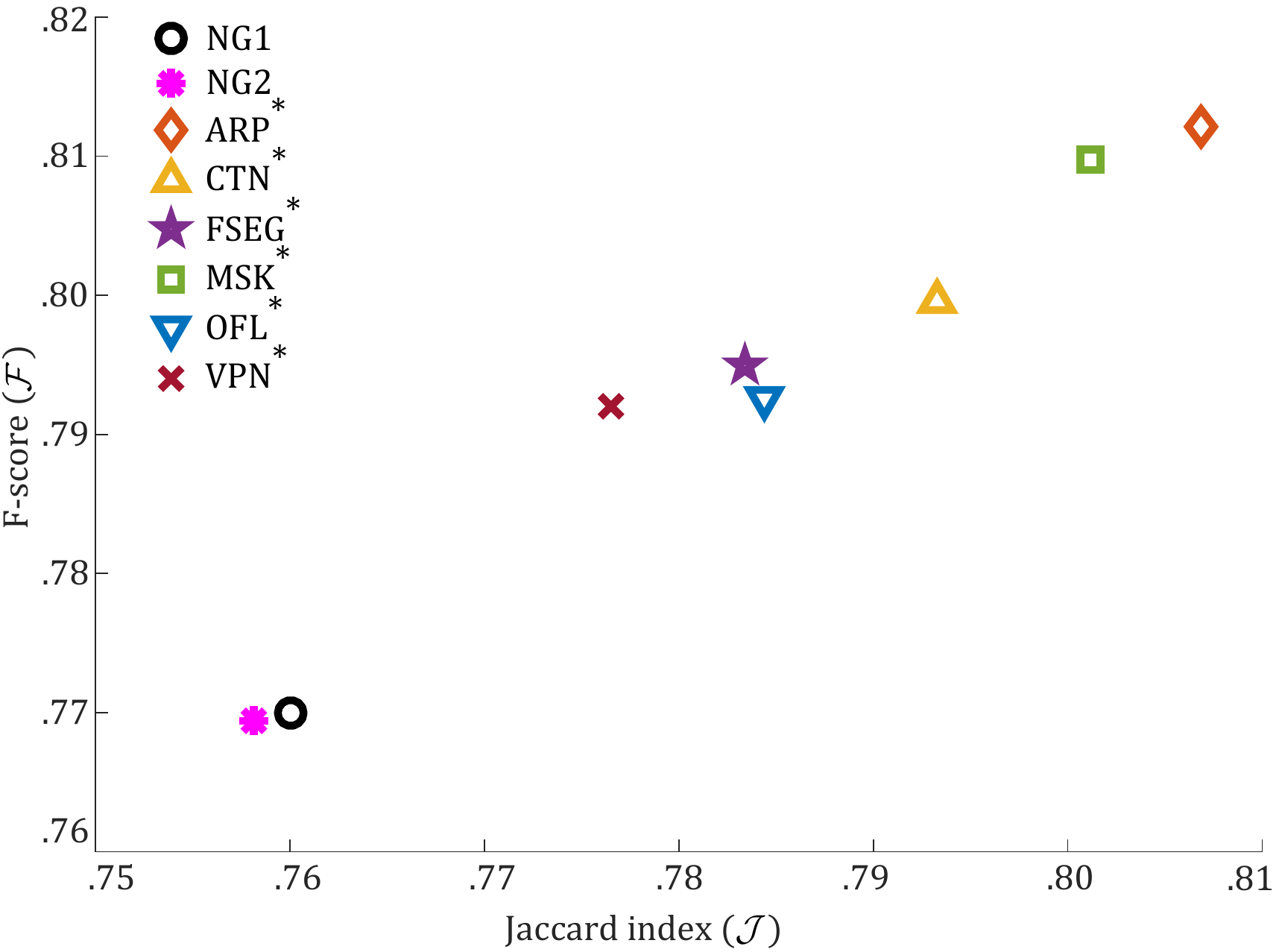}\caption{\label{fig:PerformanceVOS}Performance in DAVIS 2016 dataset of the
non-guided representations learnt (NG1 and NG2) compared to the improvements
achieved when guiding through independent foreground masks.}
\end{figure}

We compare our approach against 6 recent and top-performing VOS alternatives
(both unsupervised and semi-supervised): ARP \cite{2017_CVPR_ARP},
CTN \cite{2017_CVPR_CTN}, FSEG \cite{2017_CVPR_FSEG}, MSK \cite{2017_CVPR_MSK},
OFL \cite{2016_CVPR_OFL} and VPN \cite{2017_CVPR_VPN}. To provide
a fair comparison, we also use the segmentation masks of these alternatives
for our proposal (indicated by {*}). Table \ref{tab:PerformanceDAVIS}
presents the average performance for the 20 test sequences of DAVIS
2016 in terms of Jaccard index $\mathcal{J}$, contour F-score $\mathcal{F}$
and temporal stability $\mathcal{T}$. By comparing the independent
foreground masks and the proposed approach ({*}), we outperform all
of them in terms of $\mathcal{J}$ and $\mathcal{F}$. Regarding the
temporal stability $\mathcal{T}$, we improve for 3 of the 6 algorithms.
The reason $\mathcal{T}$ is not improved for CTN, MSK, and OFL is
that these are implicitly focused on transferring the same segmentation
frame to frame, thus resulting in an inherent temporal stability that
is difficult to improve upon from and unsupervised perspective. The
improvements achieved are due to both the potential of our learned
representations and the guiding mechanism introduced through the independent
masks. As it is not possible to separate both facts from the results
in Table \ref{tab:PerformanceDAVIS}, we perform an additional experiment
to highlight the potential of the learned representations. In particular,
we have repeated the third step of the training process (see Subsection
\ref{subsec:Implementation-details}) without making use of independent
foreground masks. We explore two non-guided alternatives, the first
one implies removing the FG-BG encoding (see Figure \ref{fig:Arquitecture}),
thus passing the representation $\mathbb{R}$ directly to the decoder
(NG1 alternative). The second alternative consists in using a dummy
foreground mask where all pixels contain foreground (i.e. $\mathbb{F}$
and $\mathbb{B}$ contain the same information), thus avoiding the
use of independent foreground masks to guide the segmentation while
using a network (NG2 alternative) with the same number of parameters
as the one proposed. Figure \ref{fig:PerformanceVOS} presents the
performance in terms of $\mathcal{J}$ and $\mathcal{F}$ for this
two non-guided alternatives (NG1 and NG2), demonstrating that non-guided
representations perform worse than the guided ones, i.e. those learnt
with separate foreground-background representation. 

\begin{figure}[t]
\centering{}\includegraphics[width=0.98\columnwidth]{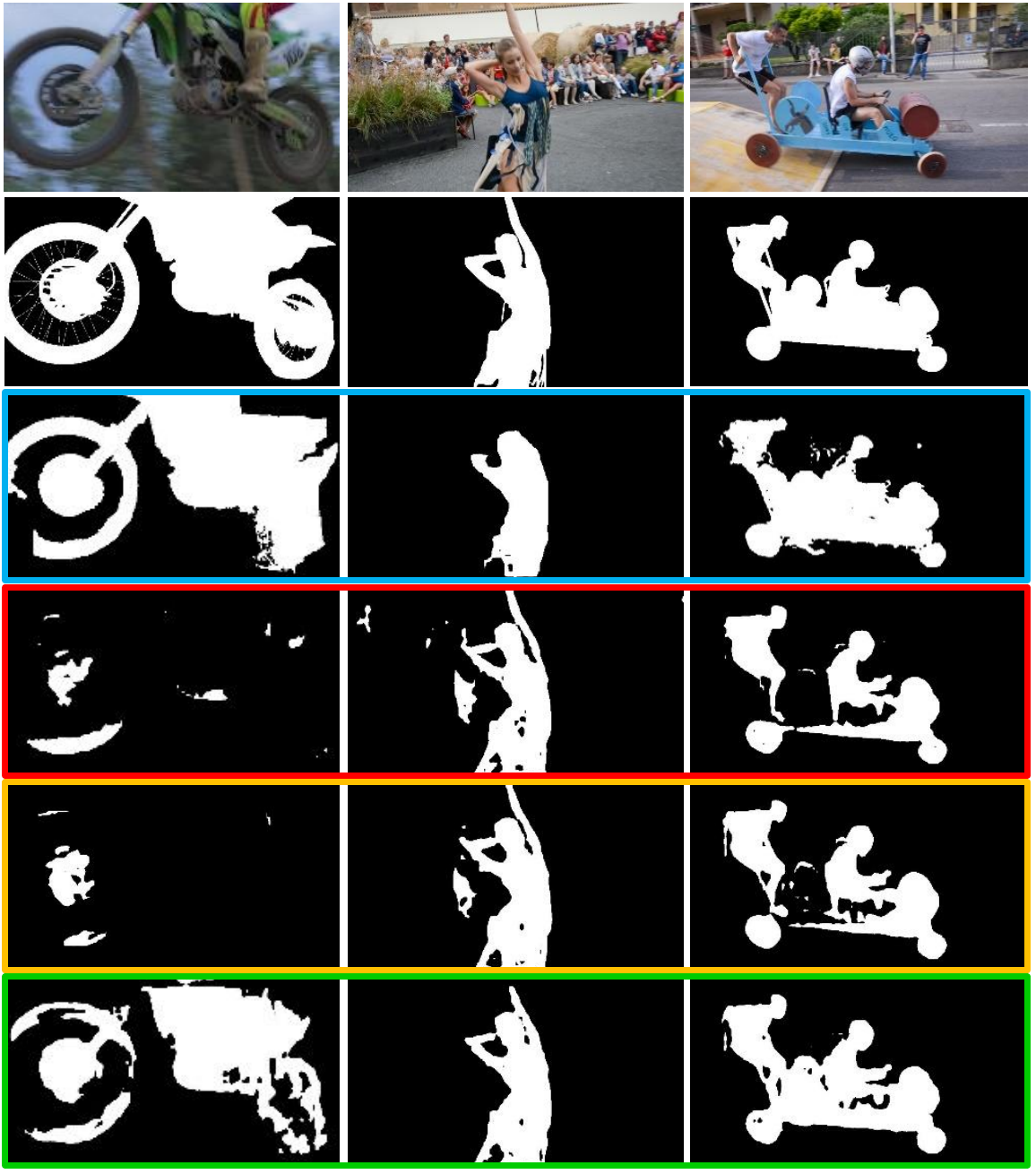}\caption{\label{fig:Foreground-segmentation-examples}Foreground segmentation
examples. From top to bottom: image to segment, ground-truth and independent
(blue), non-guided 1 (red) and 2 (orange) and proposed guided (green)
foreground segmentation masks. From left to right, examples for \emph{motocross-jump}
(ARP), \emph{dance-twirl} (FSEG) and \emph{soapbox} (VPN) video sequences.
Note that the masks with no independent guiding (red and orange) do
not depend on a particular algorithm.}
\end{figure}

Furthermore, directly comparing NG1 and NG2 performance (around $\mathcal{J}=0.76$)
with the selected alternatives in Table \ref{tab:PerformanceDAVIS}
shows that we obtain competitive results (only MSK with $\mathcal{J}=0.7955$
is better than NG1 and NG2).

Figure \ref{fig:Foreground-segmentation-examples} gives some examples
of foreground segmentations. The first column presents an example
where the proposed mask (green) is capable of solving the mistakes
of the non-guided representations (red) due to the guiding introduced
by an independent algorithm (blue). The second and third columns show
examples where the proposed approach (green) outperforms the independent
algorithm (blue) due to the guiding scheme over representations that,
without guiding (red), were lacking of object parts or introducing
false positives.

\section{Conclusions\label{sec:Conclusions}}

This paper proposes a VOS approach which takes advantage of independent
(i.e. state-of-the-art) algorithm results to guide the segmentation
process. This strategy enables learning an enhanced colour and motion
representation for VOS due to specific attention on foreground and
background classes. The experimental work validates the utility of
our approach to improve foreground segmentation performance. Future
work will explore the use of feedback strategies to induce the foreground-background
separation from the produced result and not from independent algorithms.

\section*{Acknowledgement}

This work was partially supported by the Spanish Government (MobiNetVideo
TEC2017-88169-R), ``Proyectos de cooperación interuniversitaria UAM-BANCO
SANTANDER con Europa (Red Yerun)\textquotedblright{} (2017/YERUN/02
(SOFDL) and Science Foundation Ireland (SFI) under grant numbers SFI/12/RC/2289
and SFI/15/SIRG/3283. We gratefully acknowledge the support of NVIDIA
Corporation with the donation of the Titan Xp GPU used for this research.
\bibliographystyle{IEEEtran}
\bibliography{refs}

\end{document}